\documentclass[runningheads]{llncs}
\usepackage[T1]{fontenc}
\usepackage{graphicx}
\usepackage{CJKutf8}
\usepackage{microtype}
\usepackage[main=english,vietnamese]{babel}
\usepackage[
    colorlinks=true,
    linkcolor=blue,
    anchorcolor=blue,
    citecolor=blue,
    filecolor=blue,
    urlcolor=blue
]{hyperref}
\usepackage{booktabs}
\usepackage{multirow}
\usepackage{color}

\begin{document}
\selectlanguage{english}
\title{ICDAR 2026 Competition on Writer Identification and Pen Classification from Hand-Drawn Circles}
\titlerunning{Circle\textsc{id}}
\author{
Thomas Gorges\inst{1}\orcidID{0009-0007-0573-0992} \and
Janne van der Loop\inst{2}\orcidID{0000-0001-7486-669X} \and
Lukas H{\"u}ttner\inst{1}\orcidID{0009-0001-0231-7517} \and
Linda-Sophie~Schneider\inst{1}\orcidID{0009-0002-6195-9859}\and
Fei Wu\inst{1}\orcidID{0000-0003-4196-0289} \and
Mathias Seuret\inst{1}\orcidID{0000-0001-9153-1031} \and  \\
Vincent Christlein\inst{1}\orcidID{0000-0003-0455-3799} 
}
\authorrunning{Gorges et al.}
\institute{
  Pattern Recognition Lab, Friedrich-Alexander-Universität Erlangen-Nürnberg, Germany\\
  \email{\{thomas.gorges, lukas.huettner, linda-sophie.schneider, river.wu, mathias.seuret, vincent.christlein\}@fau.de}
  \and
  Buchwissenschaft, Johannes Gutenberg-Universität Mainz, Germany\\
  \email{jannevanderloop@uni-mainz.de}
}
\maketitle              %
\setcounter{footnote}{0}
\begin{abstract}
This paper presents Circle\textsc{id}, a large-scale \textsc{icdar} 2026 competition on writer identification and pen classification from scanned hand-drawn circles.
The primary objective is to investigate how biometric writer characteristics and physical pen features naturally entangle within minimal, static traces.
Circle\textsc{id} comprises two distinct tasks: (1) open-set writer identification, requiring models to recognize known writers while explicitly rejecting unknown ones, and (2) cross-writer pen classification, evaluated across both seen and unseen writers.
Participants were provided with a new, controlled dataset of 46,155 tightly cropped circle images, digitized at 400 \textsc{dpi} and annotated for writer identity and pen type.
The dataset comprises samples from 44 known and 22 unknown writers using eight different pens.
Hosted on Kaggle as two separate tracks with public and private leaderboards, the competition provided participants with a ResNet baseline.
In total, 389 teams (436 participants) made 3,185 submissions for the pen classification task, and 113 teams (141 participants) made 1,737 submissions for the writer identification track.
The best-performing private leaderboard submissions achieved a Top-1 accuracy of 64.801\,\% for writer identification and 92.726\,\% for pen classification.
This paper details the dataset, evaluates the winning methodologies, and analyzes the impact of out-of-distribution writers on model generalization and feature disentanglement.
In this large-scale competition, Circle\textsc{id} establishes a new baseline for minimal-trace analysis.

\keywords{Document Analysis \and Writer Identification \and Pen Classification \and Open-Set Recognition.}
\end{abstract}
\section{Introduction}

Within document analysis and forensics, writer identification and pen classification are well-established research fields~\cite{bulacu2007text,BRAZ2013206}.
Studies in this field have mainly relied on information-rich inputs, such as full manuscripts, signatures, or words, which provide strong visual cues through spatial layout, allographic variations, and character shapes~\cite{bulacu2007text,hafemann2017offline,bensefia2016writer}.
However, to the best of our knowledge, identifying the writer and classifying the pen based purely on primitive shapes has not been studied.
The Circle\textsc{id} competition addresses this gap by focusing on a hand-drawn circle, thereby stripping away these information-rich details.
As a consequence, this forces machine learning models to depend exclusively on the subtle nuances of geometry and stroke texture.

A basic geometric primitive was selected for two reasons:
Even though a circle is a simple geometric shape, it still encodes kinematic behavior and instrument properties.
So, features such as starting- and endpoint locations, stroke overlap, pressure-derived ink-intensity variations, and textural ink deposition are preserved within this basic shape.
Second, biometric writer-specific and physical pen-specific features are naturally entangled within these minimal traces.
Consequently, models trained for writer identification might, therefore, unintentionally learn pen characteristics instead, and vice versa.
The goal of Circle\textsc{id} is to systematically investigate this feature entanglement and determine whether these characteristics can be isolated.

Determining the writing pen and analyzing the ink traces for characteristics, such as pressure modulation and ink deposition patterns, can provide evidence in forensic document examination~\cite{BRAZ2013206}.
So is distinguishing between minimal marks, such as a single digit `0' added to a forged check, highly challenging.
Historical document analysis benefits from Circle\textsc{id} as well, since it requires authentication of isolated symbols, annotations, or initials where comprehensive reference manuscripts are limited~\cite{boudraa2025historical}.
Furthermore, the competition results can benefit broader machine learning research, especially studying open-set recognition~\cite{geng2020recent,sun2023survey} and feature disentanglement~\cite{wang2024disentangled}.

To systematically evaluate the above described challenges, Circle\textsc{id} consists of two prediction tasks.
The first task is an open-set writer identification task, which requires models not only to recognize known writers but also to explicitly reject samples from out-of-distribution (unseen) writers.
The second task, pen classification, evaluates pen-specific features across both seen and unseen writers to determine their robustness against varying writer styles.

By defining a benchmark,
presenting quantitative participation results from the competition, and by summarizing the best-performing approaches, this paper establishes a foundational baseline for minimal-trace analysis.

\section{Competition Design}

Circle\textsc{id} was hosted as two separate competitions on the Kaggle platform.\footnote{\url{https://www.kaggle.com/competitions/icdar-2026-circleid-writer-identification}, and \url{https://www.kaggle.com/competitions/icdar-2026-circleid-pen-classification}}
Each task featured its own independent public and private leaderboards, which were evaluated using Top-1 accuracy.

\subsection{Task definitions}

\textbf{Task 1: Open-Set Writer Identification.} Participants were required to predict the identity of the writer based on images of hand-drawn circles.
The evaluation set contained samples from writers who were not available as labeled known-writer classes in the training data.
Specifically, models were required to classify these out-of-distribution samples using a dedicated ``unknown'' label class.

\noindent \textbf{Task 2: Cross-Writer Pen Classification.} With the same input images, participants were tasked to predict the specific pen used from a predefined set of classes. The test set included circles drawn by both seen and unseen writers, using only the known pens.

Performance in Task 1 required the dual ability to classify known writers and detect out-of-distribution samples.
In contrast, Task 2 required the extraction of invariant textural and morphological features that generalize across diverse writing styles.

\subsection{Timeline}

Both competitions launched on Kaggle on February 4, 2026, and ran until April 3, 2026.
During the active phase, the public leaderboards were computed using approximately 30\,\% of the held-out test data, while the remaining 70\,\% was reserved for the final private leaderboard, which was released on April 10, 2026. Participants could form teams and select up to two final submissions per task for the final evaluation.

On March 5, 2026, the evaluation set was updated to prevent the potential exploitation of a post-processing artifact that could allow an unfair advantage.
To address this, the original evaluation images were repurposed and released as additional training data, and the evaluation set generation process was revised.
By design, unknown writers in this extra dataset retained the explicit unknown writer \textsc{id}.
The original training files and formats remained unchanged.
The Kaggle leaderboards were recalculated based on this update, and all analyses in this paper are based exclusively on the corrected evaluation set.
The release of additional training data enabled participants to optionally incorporate these samples to develop more advanced methods leveraging an unseen writer set.

\subsection{Baseline}

To provide participants with a starting point, we supplied a minimal, extensible deep learning baseline for both tasks, available on GitHub\footnote{\url{https://github.com/ThomasGorges/icdar2026-circleid-baseline}} and Kaggle.\footnote{\url{https://www.kaggle.com/code/thomasgorges/circleid-writer-baseline}, and \newline\url{https://www.kaggle.com/code/thomasgorges/circleid-pen-baseline}}
We utilized a ResNet-18~\cite{he2015deep} model with weights pretrained on ImageNet~\cite{russakovsky2015imagenet} weights.

To suit the specific requirements of the Circle\textsc{id} competition, the output layer was modified to match the required number of classes.

Images were resized to $224 \times 224$ pixels, normalized using the ImageNet mean and standard deviation, and were augmented online with random rotations ($\pm 10$\textdegree).
Only samples from the original training dataset were used (without additional training).
The data was partitioned into an $80\,\%$ training set for fine-tuning the network weights, and a $20\,\%$ validation set to monitor performance and to retain the best model weights.
Network parameters were fine-tuned for ten epochs using the AdamW optimizer~\cite{loshchilov2017decoupled} with a learning rate of $3 \times 10^{-4}$ and a batch size of~128.

For the open-set writer identification task, a simple thresholding approach was used: if the maximum confidence score was below $0.9$, the sample was classified as an out-of-distribution writer.
Otherwise, the writer with the highest probability was assigned to the circle.
The network was trained exclusively on the 44 known writer classes, and out-of-distribution classes were handled via inference thresholding.
For the pen classification task, the class with the highest predicted probability was selected.

\section{Dataset}

\begin{figure}[h!]
\centering
\includegraphics[width=0.9\textwidth]{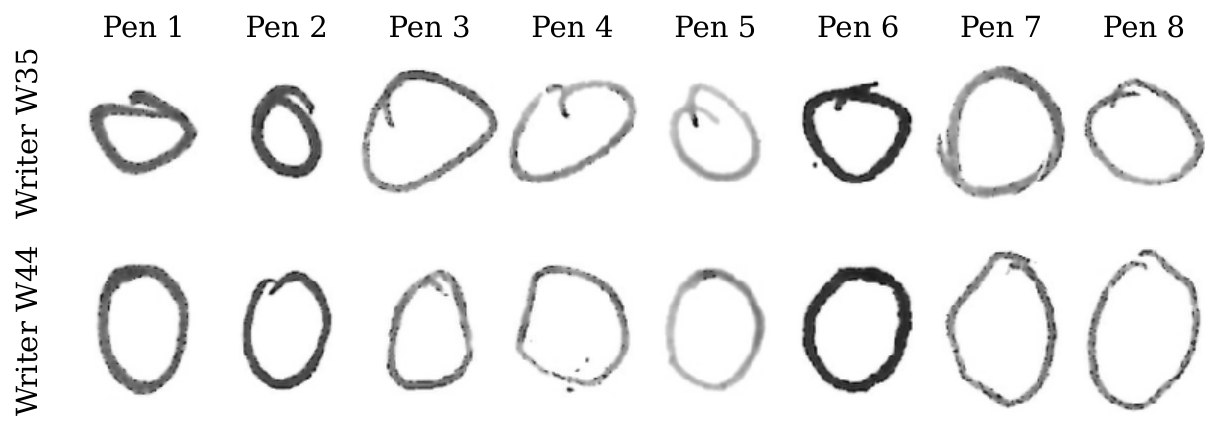}
\caption{Examples of morphological and textural variations in hand-drawn circles from randomly selected writers across the eight pens, all using black ink.} \label{fig:pen_examples}
\end{figure}

The dataset consists of a newly assembled set of hand-drawn circles, drawn under controlled conditions.
Data collection involved a cohort of 66 healthy adults from diverse cultural backgrounds, representing various age and gender groups.
Participants received standard office copy paper (A4, 80 g/m², uncoated/matte) pre-printed with a grid template.
To capture intra-writer variance, participants were instructed to draw 50 circles in a clockwise direction and 50 circles in a counterclockwise direction per pen.
Environmental variables that could affect ink flow, such as room temperature, were not monitored.

\begin{table}[h!]
\caption{Writing instruments used in the Circle\textsc{id} dataset.}
\label{tab:pens}
\centering

\begin{tabular}{cp{3cm}p{3.7cm}p{3.5cm}}
\toprule
\textbf{Pen \textsc{id}} & \textbf{Manufacturer} & \textbf{Product} & \textbf{Type} \\
\midrule
1 & \textsc{stabilo} & point 88 fine 0.4 & Fineliner \\
2 & United Office & id: 365222, black & Ballpoint \\
3 & Schneider & Slider Rave XB black & Ballpoint \\
4 & Paper Mate & EraserMate 1.0M black & Erasable ballpoint \\
5 & \textsc{pilot} & FriXion Ball black & Erasable gel rollerball \\
6 & uni-ball & Eye UB-157 black & Rollerball \\
7 & Schneider & Slider Basic XB & Ballpoint \\
8 & Schneider & K1 black & Ballpoint \\
\bottomrule
\end{tabular}
\end{table}

A curated set of eight distinct pens was preselected to evaluate the instrument classification.

All pens used black ink to minimize color-related confounding factors.
As shown in Table~\ref{tab:pens}, these instruments were chosen to cover a wide variety of writing mechanisms, tip geometries, and ink deposition properties.
Qualitative examples of the resulting visual variances between pens and writers are illustrated in Figure~\ref{fig:pen_examples}.

\subsection{Processing}

Completed templates were digitized at a resolution of 400 \textsc{dpi} using a \textsc{utax} 3262i scanner with an automatic document feeder.
To extract individual hand-drawn circles, we used an automated processing pipeline.
After manually annotating the exact positions of every cell on a blank master template, the automatic pipeline proceeds in three stages: (1) aligning each scan to this template to correct shifts, rotations, and scaling; (2) isolating and cropping the individual hand-drawn circles; (3) manually reviewing the evaluation set to remove invalid samples.

To prepare scans for alignment, images were converted to grayscale, smoothed with a Gaussian filter, and binarized using adaptive local thresholding.
Morphological closing was then applied to connect disconnected parts and strengthen the printed grid structure.
Next, we performed the template alignment in two stages: first, we estimated a coarse global alignment via a similarity transform using \textsc{sift} keypoints~\cite{lowe1999object} and \textsc{ransac}~\cite{10.1145/358669.358692}.
Second, we refined this alignment at the pixel level using the Enhanced Correlation Coefficient (\textsc{ecc}) algorithm~\cite{evangelidis2008parametric}.
For this purpose, we used a Gaussian-blurred gradient-magnitude representation of the images, combined with morphological dilation and erosion, to derive a mask that emphasizes the template's alignment structure.

Once the scans were aligned, the exact location of every cell was known.
This information was used to tightly crop the hand-drawn circles.
To do so, the handwritten ink was isolated using adaptive thresholding.
By identifying the largest connected component of foreground ink, the bounding box around the drawing was retrieved, subsequently a small margin was added, and the cell images were cropped to remove large blank spaces.

After the automated processing, the evaluation set was manually reviewed to remove low-quality samples.
Drawings where the circle overlapped with the printed grid, causing parts of the circle to be cut off, were excluded.
Furthermore, invalid drawings, improperly formed shapes, or pages with missing pen \textsc{id}s were removed. Moreover, some valid evaluation samples were discarded to artificially skew the distribution, preventing participants from exploiting distribution-aware post-processing.
To protect the privacy of the writers, their identities were anonymized.
After pre-processing, the final dataset consisted of 46,155 tightly cropped hand-drawn circle samples.

\subsection{Splits}

To prevent data leakage caused by paper-, printer-, and scan-specific artifacts, the dataset was partitioned by grouping unique writer-page-pen combinations.

Out of the total 66 writers, 44 were assigned to the training set as known writers. Samples annotated with the unknown writer class in the additional training set originated from seven other writers. One of these seven writers also appeared in the final test set and was consistently treated as unknown.
The test set consists of two parts: samples from the 16 unseen writers (Test Part A) and a held-out set of samples from seven known writers (Test Part B).
The number of total writers, unseen writers, and known writers in the test set was not announced upfront to the competitors.
Test Part A was randomly subsampled at the group level, resulting in 2,528 unseen writer samples out of 5,905 test samples, corresponding to approximately 42.81\,\% of the final test set.

For the writer identification task, all samples originating from Test Part A were mapped to the ``unknown'' class to model the open-set rejection task.
Pen labels were preserved for all samples across both tasks.
The final test set was divided into the public and private leaderboards. To guarantee isolation between the public and private sets, the division was based on the predefined groups.

During post-competition analysis, a minor annotation error was identified: one unseen writer was accidentally assigned to Test Part B instead of Part A. 
Hence, this writer was not subjected to the random group-level sub-sampling.
We remapped the ground truth labels for this unknown writer to the unknown class label, and the leaderboards were re-evaluated.
Since only the evaluation set labels were corrected, participants did not need to resubmit their predictions.

\subsection{Statistics}

\begin{figure}[tb]
\centering
\includegraphics[width=\textwidth]{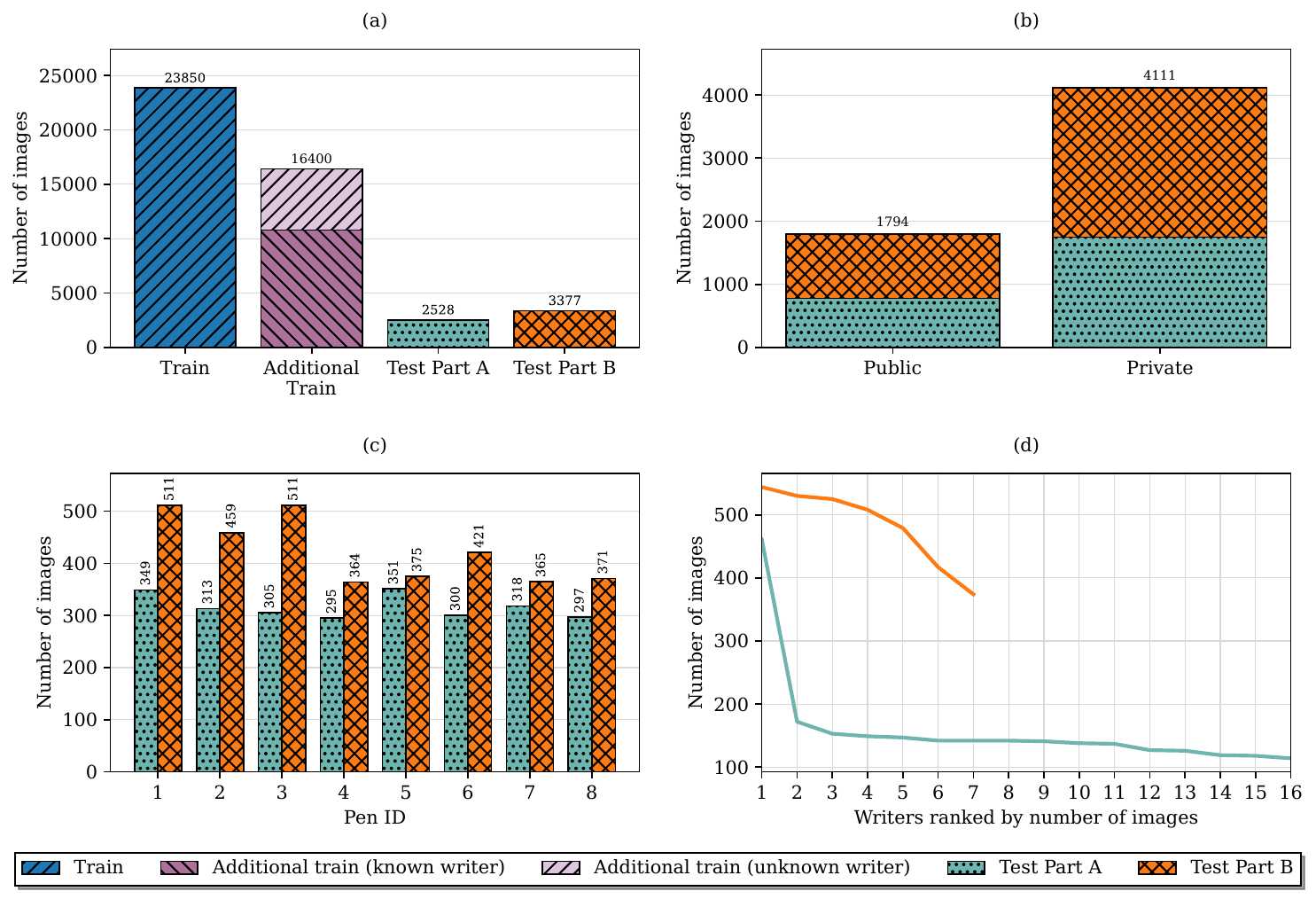}
\caption{Overview of the Circle\textsc{id} dataset splits and evaluation set distributions. \textbf{(a)}~sizes of the train set, additional train set, and evaluation set. The additional train set is subdivided into samples with known and unknown writer annotations. \textbf{(b)}~sizes of the public and private evaluation sets, further broken down by Part A and Part B. \textbf{(c)}~distribution of samples across the eight pens in the evaluation set, shown separately for Part A and Part B. \textbf{(d)}~ranked count plot showing the sample distribution across the 23 writers present in the evaluation set, shown separately for Test Parts A and B.} \label{fig:datasetOverview}
\end{figure}

As shown in Figure~\ref{fig:datasetOverview} \textbf{(a)}, the 46,155 images from 66 writers are partitioned into a primary training set (23,850 samples), an additional training set (16,400 samples), and a test set (5,905 samples).
The additional training set contains 10,800 samples from known writers and 5,600 samples from unknown writers.
Excluding the samples annotated with the unknown writer class, the combined training data contain samples from 44 known writers.
The test set is composed of 2,528 hand-drawn circles from unseen writers (Test Part A) and 3,377 circles from known writers (Test Part B).
Figure~\ref{fig:datasetOverview} \textbf{(b)} details the split of evaluation data for the competition leaderboards.
The public leaderboard contains 1,794 samples, while the private leaderboard comprises the remaining 4,111 images.
The distribution of the test set is illustrated in Figure~\ref{fig:datasetOverview}.
Sub-figure \textbf{(c)} shows the sample distribution for each pen.
In Test Part A, the images are relatively balanced across the eight pens ($316.00 \pm 22.38$).
The distribution of pens is more skewed in Test Part B ($422.12 \pm 63.96$).
This imbalance is a direct result of the group-wise sub-sampling and was introduced intentionally to prevent participants from using distribution-aware post-processing techniques.
Similarly, Figure~\ref{fig:datasetOverview} \textbf{(d)} presents ranked count plots for the writers in Test Parts A and B.
The samples from the 23 writers in the test set are not uniformly distributed.
The distribution of 15 of the 16 unseen writers was also altered.
As with the pen distributions, this artificial skewing was implemented intentionally to discourage statistical exploitation during the competition.

\section{Results}

Circle\textsc{id} attracted substantial participation for both tasks and was featured in the Kaggle Community Spotlight, ranking first and third.
In total, 436 participants across 389 teams competed in the pen classification task.
A total of 113 teams, comprising 141 participants, took part in the writer identification task.
Overall, 3,185 submissions were made for the pen classification task and 1,737 for writer identification.
Many participants also shared their approaches publicly on Kaggle: 34 notebooks were published for pen classification and 7 for writer identification.

\begin{table}[h!]
\caption{Private leaderboard containing the best 3 performing teams for the writer identification and pen classification tasks.}
\label{tab:leaderboard_combined}
\centering

\begin{minipage}[t]{0.48\textwidth}
\centering
\begin{tabular}{clcr}
\toprule
\multicolumn{4}{c}{\textbf{Writer identification}} \\
\midrule
\textbf{Place} & \textbf{Team name} & \shortstack{\textbf{Top-1}\\\textbf{acc.}} & \shortstack{\textbf{Rank}\\\textbf{shift}} \\
\midrule
1 & \begin{CJK*}{UTF8}{gbsn}I-Signing亲笔签\end{CJK*} & 64.801\,\% & 0 \\
2 & mpwi0410 & 53.612\,\% & +1 \\
3 & busy & 51.885\,\% & +1 \\
\bottomrule
\end{tabular}
\end{minipage}
\hfill
\begin{minipage}[t]{0.48\textwidth}
\centering
\begin{tabular}{clcr}
\toprule
\multicolumn{4}{c}{\textbf{Pen classification}} \\
\midrule
\textbf{Place} & \textbf{Team name} & \shortstack{\textbf{Top-1}\\\textbf{acc.}} & \shortstack{\textbf{Rank}\\\textbf{shift}} \\
\midrule
1 & \foreignlanguage{vietnamese}{\textsc{uit}-PhởMậu} & 92.726\,\% & +8 \\
1 & \textsc{iitp\_cv\_14} & 92.726\,\% & +17 \\
3 & Yoshi187 Jhoshua & 92.629\,\% & +9 \\
\bottomrule
\end{tabular}
\end{minipage}

\end{table}

\subsection{Challenge winners}

The top-ranked team on the private leaderboard for writer identification was \begin{CJK*}{UTF8}{gbsn}\textit{I-Signing亲笔签}\end{CJK*} (Xunhui Qin, Zhonghao Shen, Du Zhou), achieving a Top-1 accuracy of $64.801\,\%$.
This team also ranked first on the public leaderboard.
The second-place team, \textit{mpwi0410} (Marco Peer), achieved $53.612\,\%$, followed by \textit{busy} in third place with $51.885\,\%$ accuracy.
For comparison, our thresholding-based benchmark achieved a private leaderboard score of $38.384\,\%$ and a public score of $42.363\,\%$.

For the pen classification task, two teams achieved the same Top-1 accuracy on the private leaderboard: \foreignlanguage{vietnamese}{\textit{\textsc{uit}-PhởMậu}} (Huy Nguyen Pham Gia, Dang Ngo Viet Tue, Hien Pham Duy, Bao Ta Cao Nguyen) and \textit{\textsc{iitp\_cv\_14}} (Dheeraj Banavath, Pitta Tanuj, Marella Sathvik, Kumari Priya, Bibek Das), both reaching first place with $92.726\,\%$ accuracy.
Third place was achieved by \textit{Yoshi187 Jhoshua} with $92.629\,\%$.
Here, our benchmark achieved $85.429\,\%$ on the private leaderboard and $80.546\,\%$ as public score.

\subsection{Winner analysis}

\begin{figure}[h!]
\centering
\includegraphics[width=\textwidth]{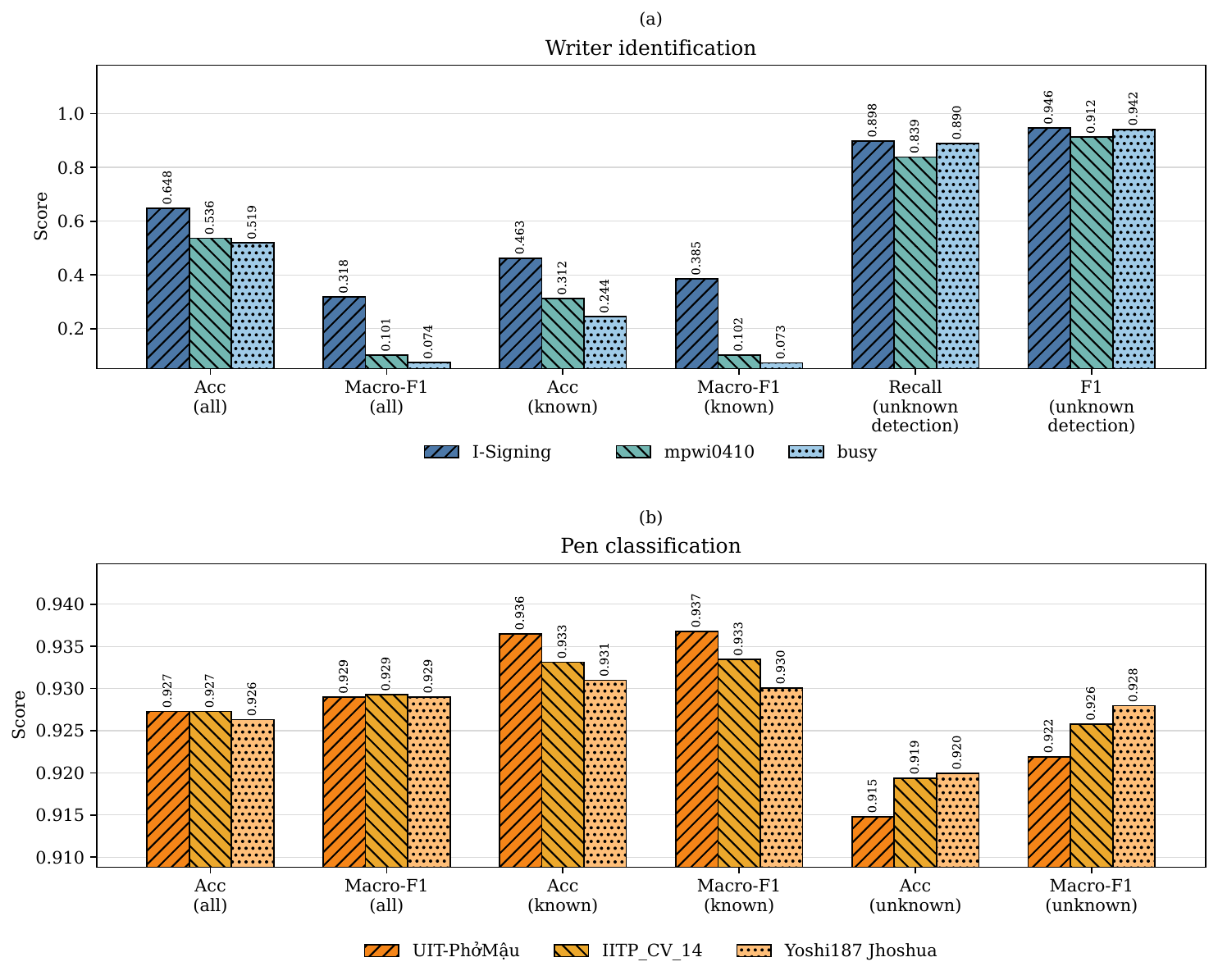}
\caption{Performance analysis of the top-3 teams on different subsets of the private leaderboard data. \textbf{(a)} writer identification task. Accuracies represent Top-1 accuracies, F1 scores for all categories, and macro-F1 scores for known categories. Recall and F1 for unknown writers are recall and binary F1 for unknown-writer detection on the unknown-writer subset. \textbf{(b)} pen classification task. Top-1 accuracies and macro-F1 scores are shown.} \label{fig:top3_compare}
\end{figure}

A more detailed comparison of the top-3 teams is shown in Figure~\ref{fig:top3_compare} for writer identification in \textbf{(a)} and pen classification in \textbf{(b)}.
For writer identification, the winning team achieved a substantially higher macro-F1 on the known-writer subset ($0.385$) than the second-place team ($0.102$), which contributed strongly to its superior overall performance.
In contrast, unknown-writer detection performance was relatively similar across the top-3 teams.

For pen classification, a different pattern is visible.
Performance on samples from known writers was consistently higher than on samples from unknown writers.
At the same time, teams that performed particularly well on the known-writer subset did not necessarily achieve the strongest results on the unknown-writer subset, suggesting a trade-off in generalization across writer conditions.

\subsection{Winning strategies}

To understand the top-ranked submissions, the main methodologies reported by the top teams are summarized below.

\paragraph{Writer identification (\begin{CJK*}{UTF8}{gbsn}I-Signing亲笔签\end{CJK*}: Xunhui Qin, Zhonghao Shen, Du Zhou).}

The method relied on strong data augmentation, combining geometric operations such as cropping, rotation, and affine transforms with color jitter and a dedicated novel stroke-width simulation scheme to improve robustness to handwriting variation.
The core model consisted of three complementary branches: a sequential branch that represented elliptical trajectories as $0$--$360^\circ$  pseudo-sequences and modeled them with a \textsc{1d-cnn}, and two image-based branches that processed skeletonized inputs and raw color images separately using weight-unshared ConvNeXt backbones~\cite{liu2022convnet}.
The team trained the model using a cross-entropy classification loss combined with a contrastive metric-learning loss, aiming to separate the known writers while keeping feature clusters compact enough to support rejection of unknown writers.
During inference, the final decision was obtained by combining class logits with distances in the learned feature space of the test set, exploiting both classification-based and representation-based information.
Based on behavior observed across multiple submissions, the team found that the model predicted known classes too often on the test data compared to the validation split.
To counter this effect, they applied a fixed-ratio rejection rule: samples were ordered by the Top-1 logit of the known classes, and the bottom $65\,\%$ with the lowest confidence were assigned the unknown writer class.

\paragraph{Writer identification (mpwi0410: Marco Peer).}

The method used an ensemble of five ResNet50~\cite{he2015deep} models with Generalized Max Pooling~\cite{murray2014generalized}, trained in a metric-learning setting with Multi-Similarity Loss~\cite{wang2019multi}.
Unknown writers were excluded during training, so the learned embedding space focused on separating known writer identities while remaining suitable for later rejection of unseen classes.
The feature embeddings from the five models were concatenated to form a single representation for each sample.
For inference, the team trained Exemplar \textsc{svm}s~\cite{malisiewicz2011ensemble} on these concatenated embeddings and determined a separate distance threshold for each writer.
These thresholds were estimated using samples labeled as unknown writers from the additional training data and were used to reject test samples that did not closely enough match any known writer.

\paragraph{Pen classification (\foreignlanguage{vietnamese}{\textsc{uit}-PhởMậu}: Huy Nguyen Pham Gia, Dang Ngo Viet Tue, Hien Pham Duy, Bao Ta Cao Nguyen).}

The method combined two complementary processing streams for handwriting analysis.
The first stream operated directly on raw handwriting images and was trained with extensive data augmentation together with hard-pair oversampling to emphasize challenging samples.
The second stream extracted handcrafted physical descriptors designed to capture frequency structure, micro-roughness, and edge-tracking information.
These inputs were processed by a dual-stream encoder, using a \textsc{dino}v3~\cite{simeoni2025dinov3} \mbox{ConvNeXt-B}~\cite{liu2022convnet} model with attention for the image stream and a multilayer perceptron (\textsc{mlp}) for the physical-feature stream, after which both representations were concatenated.
Optimization was performed with AdamW~\cite{loshchilov2017decoupled}, a multi-task objective combining ArcFace~\cite{deng2019arcface}, and cross-entropy loss, aiming to improve both discriminative feature learning and classification accuracy.
At inference, the final prediction was obtained through a refined inference pipeline.
The team applied test-time augmentation to stabilize predictions across transformed versions of the same sample, combined outputs from a 2-fold ensemble to reduce variance, and used logit calibration to improve confidence estimation.
The calibrated ensemble output was then used as the final prediction.

\paragraph{Pen classification (\textsc{ittp}\_\textsc{cv}\_14: Dheeraj Banavath, Pitta Tanuj, Marella Sathvik, Kumari Priya, Bibek Das).}

The team combined the original training and additional training data.
Inputs in \textsc{rgb} format were resized to a fixed resolution and normalized with standard ImageNet~\cite{russakovsky2015imagenet} statistics.
The approach combined two modern \textsc{cnn} backbones, ConvNeXt-Tiny~\cite{liu2022convnet} and EfficientNetV2-S~\cite{tan2021efficientnetv2}.
Both models were initialized from ImageNet pretraining to support fine-grained texture modeling.
Training was carried out in multiple stages.
In the first stage, the models were optimized on labeled data using cross-entropy loss and label smoothing, with strong augmentations including rotation, translation, scale jitter, and brightness and contrast changes, along with MixUp~\cite{zhang2017mixup} and CutMix~\cite{yun2019cutmix}.
In the second stage, predictions on the test set were used as pseudo-labels, and the models were retrained on the union of original and pseudo-labeled samples.
This multi-stage procedure helped the models capture subtle pen-specific cues more effectively.
During inference, the team applied an extended test-time augmentation strategy, repeatedly evaluating each image under random transformations such as rotation, cropping, flipping, and mild color jitter. 
These resulting predictions were averaged for stability.
The final output was produced with a weighted ensemble of several trained models, where each model contributed in proportion to its validation accuracy rather than through uniform averaging.
According to the team, this weighting scheme further improved the overall performance.

\section{Analysis}

To better understand the observed performance, several additional analyses were conducted. Specifically, we evaluate the stability in the leaderboard rankings, analyze task-specific performance across known and unknown writers, and perform a qualitative error analysis. These evaluations provide insight into model generalization and the limits of feature disentanglement.

\subsection{Leaderboard stability}

\begin{figure}[tb]
\centering
\includegraphics[width=\textwidth]{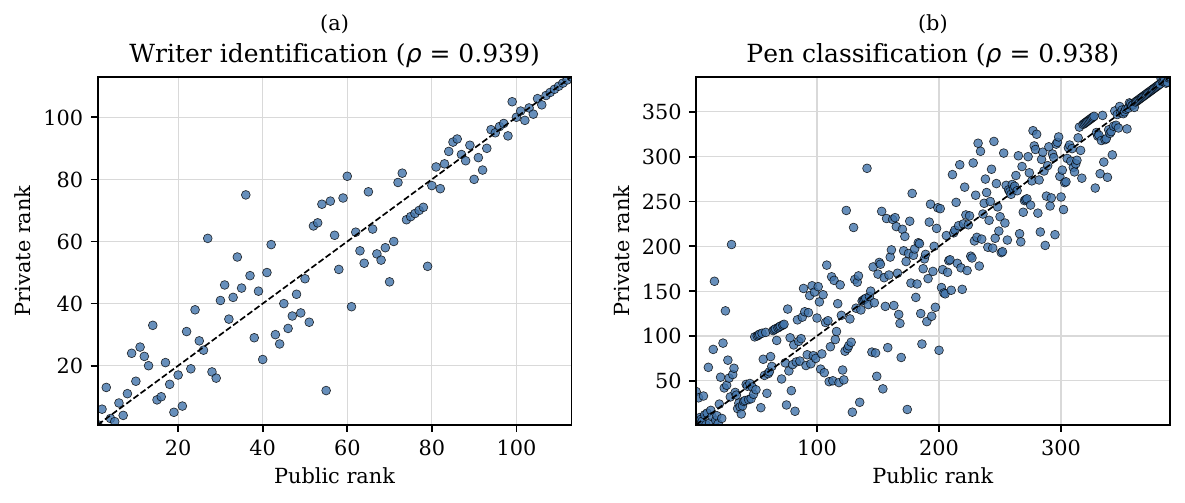}
\caption{Rank-shift analysis between the public and private leaderboards. \textbf{(a)} writer identification task. \textbf{(b)} pen classification task.} \label{fig:rank_shift}
\end{figure}

We analyzed the rank shifts between the public and private leaderboards for both tasks.
Figure~\ref{fig:rank_shift} shows the rank changes for writer identification in \textbf{(a)} and pen classification in \textbf{(b)}.
The Spearman rank correlations are high for both tasks, with $\rho = 0.939$ for writer identification and $\rho = 0.938$ for pen classification.
This indicates that the rankings are relatively stable and suggests that the public and private leaderboard sets follow similar distributions.

\begin{figure}[tb]
\centering
\includegraphics[width=\textwidth]{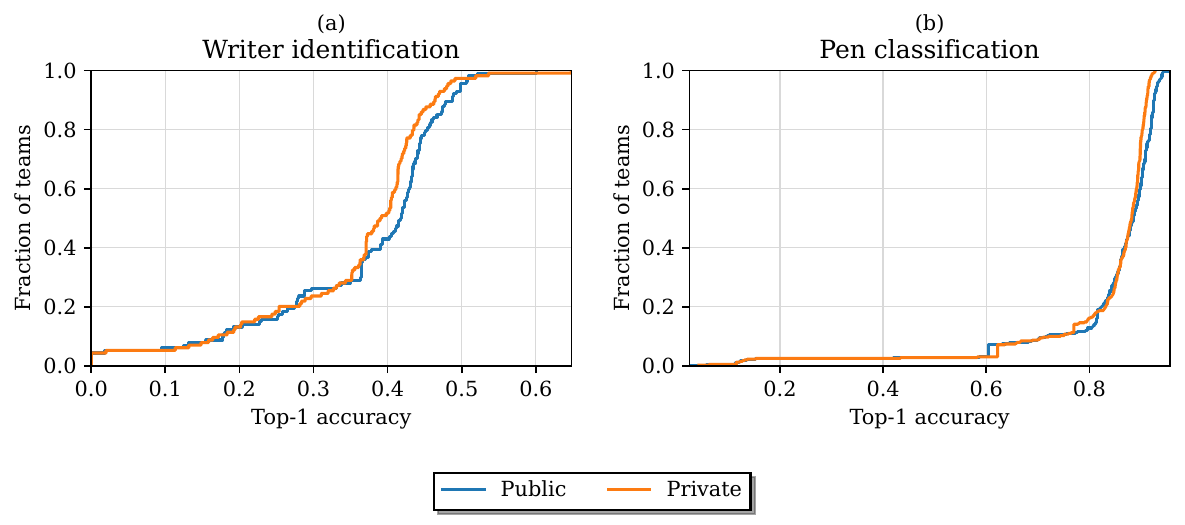}
\caption{Empirical cumulative distribution functions of Top-1 accuracy for both tasks, shown separately for the public and private leaderboards. \textbf{(a)} writer identification leaderboard. \textbf{(b)} pen classification leaderboard.} \label{fig:leaderboard_ecdf}
\end{figure}

We also examined how strongly the achieved performance shifted between the public and private leaderboards. Figure~\ref{fig:leaderboard_ecdf} shows the empirical cumulative distribution functions for \textbf{(a)} writer identification and \textbf{(b)} pen classification.
For writer identification, the median Top-1 accuracy was $41.806\,\%$ on the public leaderboard and $39.078\,\%$ on the private leaderboard.
For pen classification, the median Top-1 accuracy was substantially higher, with $88.573\,\%$ on the public leaderboard and $88.227\,\%$ on the private leaderboard.
These results indicate that pen classification is not as challenging as writer identification, while the shift between public and private performance remains limited for both tasks.

\subsection{Writer identification}

\begin{figure}[tb]
\centering
\includegraphics[width=\textwidth]{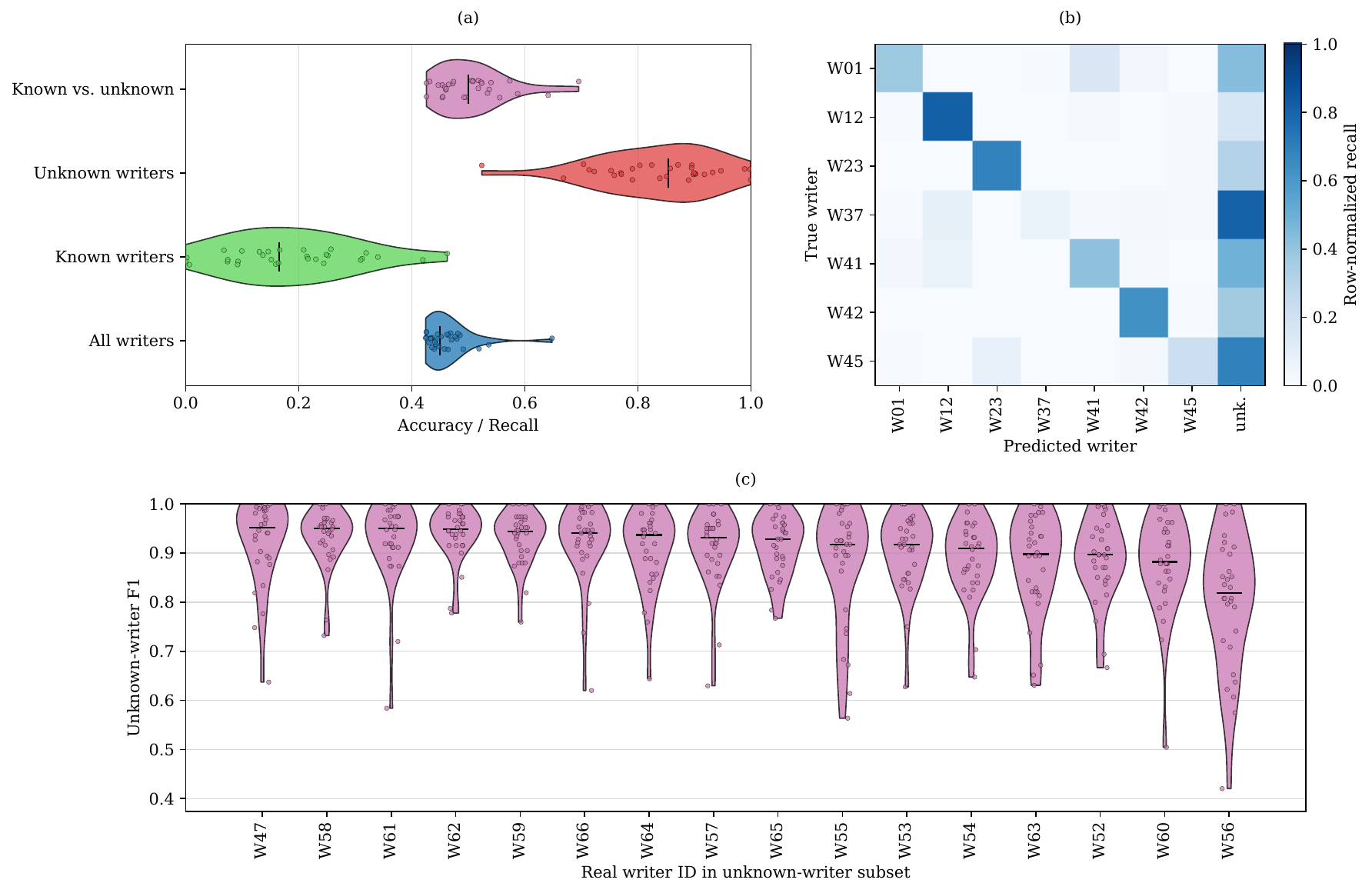}
\caption{Writer identification top-30 team performance on the private leaderboard evaluation set. \textbf{(a)} violin plots showing the distribution across the all-writer Top-1 accuracy, known-writer Top-1 accuracy, and unknown-writer recall, and binary known vs. unknown accuracy. In this binary setting, a prediction counts as correct when the model correctly distinguishes between any known writer and the unknown class, independent of whether the specific known writer identity is correct. \textbf{(b)} confusion matrix on the known-writer subset. \textbf{(c)} performance on the unknown-writer subset grouped by the underlying real writer identity. Binary F1 is shown.} \label{fig:writerResults}
\end{figure}

Figure~\ref{fig:writerResults} \textbf{(a)} provides a more detailed analysis of the top-30 private-leaderboard teams for writer identification across three subsets: all writers, known writers, and unknown writers.
A counterintuitive observation is that detecting whether a sample originates from an unknown writer appears easier than identifying the correct writer among known ones.
The results on the unknown-writer subset show comparatively strong recall and detection F1 scores, whereas performance on the all-writer and known-writer subsets remains substantially lower.
This suggests that a circle contains enough information to detect an unknown writer, but not enough features to accurately distinguish between known writers.

A confusion matrix for the known-writer subset is shown in Figure~\ref{fig:writerResults} \textbf{(b)}.
Rather than confusing known writers, models tend to misclassify samples as unknown.
This suggests that the models either identify the correct known writer or default to the unknown class.

Figure~\ref{fig:writerResults} \textbf{(c)} shows performance on the unknown-writer subset grouped by the underlying real writer identity.
The binary F1 for unknown-writer detection is shown.
This analysis was conducted to determine whether unknown-writer detection is stable across different writers or varies substantially between them.
For almost all unseen writers, both measures remain relatively stable, with only one writer showing a stronger deviation.
Overall, these results indicate that out-of-distribution samples from unknown writers can be reliably detected. In a forensic context, this implies that minimal fraudulent additions, such as a forged `0' added to a check, could be successfully flagged as an anomaly.

\subsection{Pen classification}

\begin{figure}[tb]
\centering
\includegraphics[width=\textwidth]{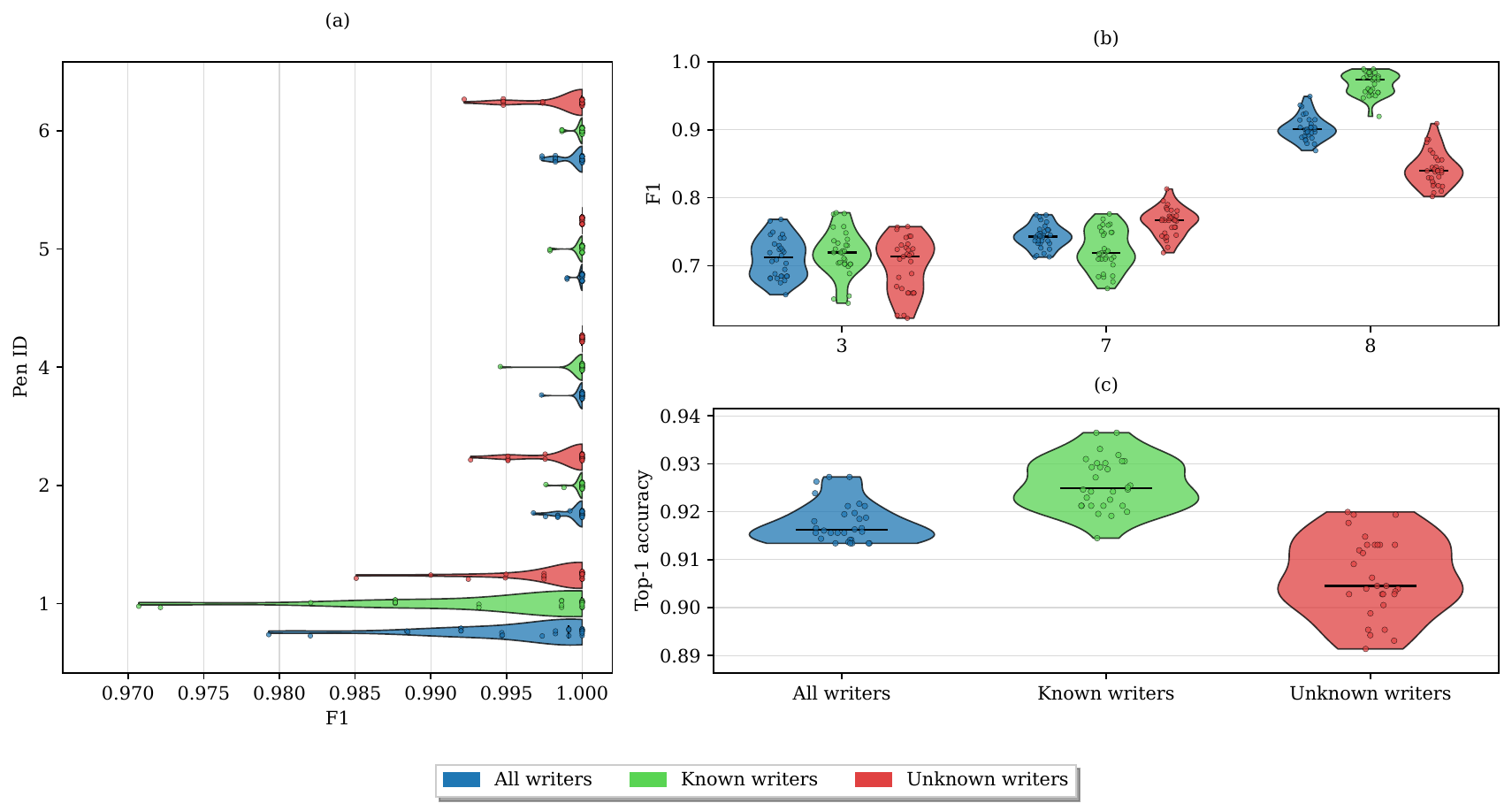}
\caption{Pen classification top-30 team performance on the private leaderboard evaluation set. \textbf{(a)} per-pen F1 distribution for pens 1, 2, 4, 5, and 6, shown separately for all, known, and unknown writers. \textbf{(b)} per-pen F1 distribution for pens 3, 7, and 8 under the same subset split. \textbf{(c)} top-1 accuracy distribution of teams across all, known, and unknown writer subsets.} \label{fig:penResults}
\end{figure}

For the pen classification task, we analyzed which pens could be reliably classified. Figure~\ref{fig:penResults} \textbf{(a)} \& \textbf{(b)} shows per-pen performance for the top-30 teams, also split into the all-writer, known-writer, and unknown-writer subsets.
The models recognize pens 1, 2, 4, 5, and 6 consistently with strong performance, whereas pens 3, 7, and~8 prove more challenging to classify.
These last three pens are all ballpoint pens produced by Schneider.
Although Pen 2 is also a ballpoint pen, it is made by a different manufacturer.
We therefore hypothesize that the tips and resulting stroke characteristics of Pens 3, 7, and 8 are visually more similar, making them harder to distinguish.
The results also suggest that including samples from the same writers during training is beneficial, yielding a slight performance gain.
Because pen classification shows strong performance even on unseen writers, we conclude that pen-specific characteristics can be successfully extracted independent of writer characteristics, demonstrating feature disentanglement.

This trend becomes clearer when aggregating results across all pens, as shown in Figure~\ref{fig:penResults} \textbf{(c)}. 
While writer overlap between training and evaluation is not strictly necessary for strong pen classification performance, it is beneficial.

\subsection{Error analysis}

\begin{figure}[h!]
\centering
\includegraphics[width=\textwidth]{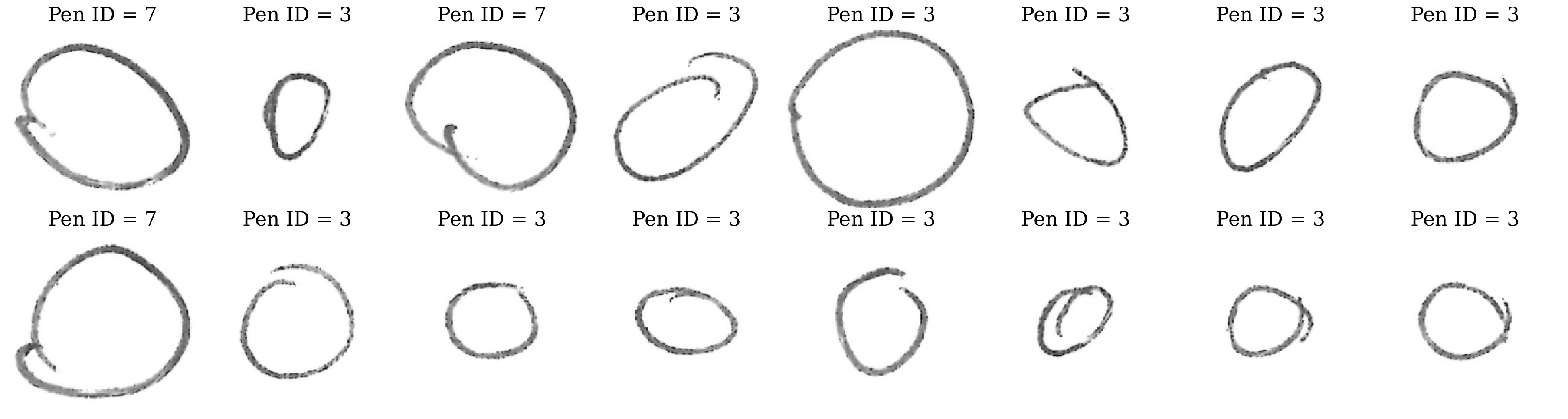}
\caption{Samples from the private leaderboard set that were most frequently misclassified by the top-30 teams in the pen classification task.
Ground truth of all samples is shown.} \label{fig:error_analysis}
\end{figure}

Figure~\ref{fig:error_analysis} visualizes the 24 samples most frequently misclassified by the top-30 teams in the pen classification task.
Most of these samples belong to Pens 3 and~7, both Schneider ballpoint pens.
Visual inspection suggests that their stroke textures and line characteristics are highly similar, which may explain their lower classification performance compared with those of the other pens.

\section{Discussion}

Pen classification proved easier, while writer identification from isolated circles was considerably more difficult.
This difference suggests that pen-specific stroke texture and material properties can be captured more readily from circles than writer-specific features.

Within the writer identification task, the results revealed a counterintuitive finding: detecting samples from unseen writers was easier than assigning a sample to the correct known writer.
This suggests that circles may contain sufficient information to determine whether a sample originates from outside the training set, but not enough writer-specific detail for reliable identification of a known individual.
More broadly, these findings indicate that writer identification from very simple static traces is a challenging problem.
Therefore, further research in this direction is needed.

For pen classification, performance was strong for most pens, but several ballpoint pens remained difficult to separate.
In particular, the recurring confusion among specific Schneider ballpoint pens suggests that similar stroke textures and line characteristics can limit discriminability, even for high-performing methods.
This observation is consistent with the qualitative error analysis.
Moreover, our analyses indicate that writer-specific features do not significantly affect pen classification performance, and that writer- and pen-specific features can be disentangled to achieve a generalizable solution.

At the same time, the study has limitations.
All samples were collected on one paper type, and environmental factors that may affect ink flow were not measured.
Future work could investigate more varied recording conditions, and explainable \textsc{ai} to better understand important visual features.

\section{Conclusion}

In this paper, Circle\textsc{id}, the \textsc{icdar} 2026 competition on writer identification and pen classification from hand-drawn circles, was presented.
The competition focused on a very primitive shape and provided a new benchmark for studying writer and pen features under limited conditions.
The results show that detecting unknown writers is easier than identifying known writers, while pen classification is highly reliable for most pens.
However, writer identification from minimal traces remains difficult and requires more research.

\begin{credits}
\subsubsection{\ackname} We thank all participants whose time and effort made the Circle\textsc{id} competition possible.
We are grateful to Ambros Marzetta and Quan Nguyen (Viettel) for reporting a post-processing artifact that could compromise evaluation integrity, and to Kaggle for hosting and featuring Circle\textsc{id} in the community spotlight.

We are grateful to the following participants for providing insights about their approaches (in alphabetical order): Alexey Senchenko, Ambros Marzetta, Andrii Grygoriev, Ayako Sato, Bao Ta Cao Nguyen, Bibek Das, Bui Huynh Tay, CHEN Meng, Dang Ngo Viet Tue, Dheeraj Banavath, Dmytro Kozii, Du Zhou, Eiki Murata, Gregor Brunner, Hien Pham Duy, Huy Nguyen Pham Gia, \foreignlanguage{vietnamese}{Huỳnh Trung Cường}, Jiahao Liu, Kumari Priya, Leander Schnurrer, Manan Bhansali, Marco Peer, Marella Sathvik, Md Raihan, Mikhail Kuznetsov, Nguyen Thanh Son, \foreignlanguage{vietnamese}{Nguyễn Khánh Tài}, Pham Nguyen Gia Huy, Phani Srikanth, Pitta Tanuj, Quan Nguyen, Rento Yamaguchi, Tashin Ahmed, Viktor Zaytsev, \foreignlanguage{vietnamese}{Vũ Gia Bảo}, Xunhui Qin, Zhonghao Shen.

The authors gratefully acknowledge the scientific support and \textsc{hpc} resources provided by the Erlangen National High Performance Computing Center (\textsc{nhr@fau}) of the Friedrich-Alexander-Universität Erlangen-Nürnberg (\textsc{fau}). The hardware is funded by the German Research Foundation (\textsc{dfg}).

\subsubsection{\contriblistname} \textsc{t.g.} conceived, designed, managed, and hosted the competition, curated and processed the dataset, conducted the analysis, and drafted the manuscript. \textsc{j.v.d.l.}, \textsc{f.w.}, \textsc{m.s.}, and \textsc{v.c.} provided feedback for the challenge design. All authors reviewed and edited the manuscript.

\subsubsection{\discintname}
The authors have no competing interests to declare that are relevant to the content of this article.
\end{credits}
\bibliographystyle{splncs04}
\bibliography{references}
\end{document}